\documentclass{article} 
\usepackage{iclr2021_conference,times}

\usepackage{graphicx}
\graphicspath{ {./pics/} }

\usepackage{amsmath,amsfonts,bm}

\def\eqref#1{equation~\ref{#1}}

\def\1{\bm{1}}

\def\rvc{{\mathbf{c}}}

\def\rvx{{\mathbf{x}}}

\def\rvz{{\mathbf{z}}}

\def\va{{\bm{a}}}

\def\vc{{\bm{c}}}

\def\ve{{\bm{e}}}

\def\vx{{\bm{x}}}

\def\vz{{\bm{z}}}

\def\evc{{c}}

\def\mA{{\bm{A}}}

\def\mE{{\bm{E}}}

\def\mZ{{\bm{Z}}}

\DeclareMathAlphabet{\mathsfit}{\encodingdefault}{\sfdefault}{m}{sl}
\SetMathAlphabet{\mathsfit}{bold}{\encodingdefault}{\sfdefault}{bx}{n}

\def\gD{{\mathcal{D}}}
\def\gE{{\mathcal{E}}}

\def\gP{{\mathcal{P}}}
\def\gQ{{\mathcal{Q}}}

\def\gT{{\mathcal{T}}}

\def\sC{{\mathbb{C}}}

\def\sH{{\mathbb{H}}}

\def\sR{{\mathbb{R}}}

\def\sX{{\mathbb{X}}}

\def\sZ{{\mathbb{Z}}}

\newcommand{\E}{\mathbb{E}}
\newcommand{\Ls}{\mathcal{L}}

\newcommand{\reg}{\lambda}

\newcommand{\softmax}{\mathrm{softmax}}

\newcommand{\KL}{D_{\mathrm{KL}}}

\DeclareMathOperator*{\argmin}{arg\,min}

\renewcommand{\rvx}{\mathtt{x}}
\renewcommand{\rvc}{\mathtt{c}}
\renewcommand{\rvz}{\mathtt{z}}
\newcommand{\dd}{\, \textnormal{d}}
\newcommand{\Lsh}{\widehat{\Ls}}
\newcommand{\pc}{p_c}
\newcommand{\qc}{q_c}
\newcommand{\mux}{\mu_x}
\newcommand{\muc}{\mu_c}
\newcommand{\Hc}{\sH_{\muc}}
\newcommand{\cHc}{\sH_{\muc|\qc}}
\newcommand{\gEt}{\gE_\theta}
\newcommand{\gQE}{\gQ_\mE}
\newcommand{\gDp}{\gD_\phi}
\newcommand{\gTEt}{\gT_{\mE, \theta}}
\newcommand{\vzb}{\bar{\vz}}
\newcommand{\vzh}{\hat{\vz}}
\newcommand{\vzt}{\tilde{\vz}}
\newcommand{\vxh}{\hat{\vx}}
\newcommand{\gPp}{\gP_{\psi}}

\usepackage{hyperref}
\usepackage{url}

\title{Learned transform compression with optimized entropy encoding.}

\author{Magda Gregorov\'a \& Marc Desaules \& Alexandros Kalousis \\
Geneva School of Business Administration, \\
HES-SO University of Applied Sciences of Western Switzerland \\
\texttt{\{name.surname\}@hesge.ch} \\
}

\iclrfinalcopy 
\begin{document}

\maketitle

\begin{abstract}
We consider the problem of learned transform compression where we learn both, the transform as well as the probability distribution over the discrete codes.
We utilize a soft relaxation of the quantization operation to allow for back-propagation of gradients and employ vector (rather than scalar) quantization of the latent codes.
Furthermore, we apply similar relaxation in the code probability assignments enabling direct optimization of the code entropy.
To the best of our knowledge, this approach is completely novel.
We conduct a set of proof-of concept experiments confirming the potency of our approaches.
\end{abstract}

\section{Introduction}\label{sec:Intro}

We consider the problem of compressing data $\vx \in \sX$ sampled i.i.d. according to some unknown probability measure (distribution) $\vx \sim \mux$.
We take the standard transform coding \citep{sayoodIntroductionDataCompression2012} approach where we first transform the data $\vx$ by a learned non-linear function, an encoder $\gEt : \sX \to \sZ$, into some latent representation $\vz \in \sZ$.
We then quantize the transformed data $\vz = \gEt(\vx)$ using a quantization function $\gQE : \sZ \to \sC$ parametrized by learned embeddings $\mE$ (codebook/ dictionary) so that the discrete codes composed of indexes of the embedding vectors $\vc = \gQE(\vz)$ can be compressed by a lossless entropy encoding and transmitted.
The received and losslessly decoded integer codes are then used to index the embedding vectors and dequantized back to the latent space $\overline{\gQE} : \sC \to \widehat{\sZ} \subset \sZ$ introducing a distortion due to mapping the codes only to the discrete subset $\widehat{\sZ} \subset \sZ$ corresponding to the quantization embeddings.
The dequantized data $\vzh = \overline{\gQE}(\vc)$ are then decoded by a learned non-linear decoder $\gDp : \widehat{\sZ} \to \sX$ to obtain the reconstructions $\vxh = \gDp(\vzh)$.

Our aim is to learn the transform (encoder/decoder) as well as the quantization so as to minimize the expected distortion $\E_{\mux} d(\rvx, \hat{\rvx})$\footnote{We use $\rvx, \rvz, \rvc$ for random variables and $\vx, \vz, \vc$ for their realizations.} while, at the same time, minimizing the expected number of bits transmitted (the rate) when passing on the discrete codes $\E_{\muc} l(\rvc)$, where $l$ is the length of the bit-encoding.
The two competing objectives are controlled via a hyper-parameter $\reg$
\begin{align}\label{eq:trade_off}
\Ls := \underbrace{\E_{\mux} d(\rvx, \hat{\rvx})}_{distortion} + \reg \underbrace{\E_{\muc} l(\rvc)}_{rate} \enspace .
\end{align}
The optimal length of encoding a symbol $\vc \sim \muc$ is determined by Shannon's self information\footnote{When considering bit-encoding, the $\log$ should be with base 2 instead of the natural base for nats.} $i_c(\vc) = - \log \pc(\vc)$  \citep{coverElementsInformationTheory2006}, where $\pc$ is the discrete probability mass\footnote{The probability mass $\pc$ is the probability density function of $\muc$ with respect to the counting measure $\muc(\rvc \in \mA) = \int_A \pc \dd \# = \sum_{\va \in \mA} \pc(\va)$, such that $\pc(\va) = \pc(\rvc = \va) = \muc(\rvc = \va)$.} of the distribution $\muc$.
Consequently, the expected optimal description length for the discrete code $\rvc$ can be bounded by its entropy $\Hc(\rvc) = - \E_{\muc} \log \pc(\rvc)$ as $\Hc(\rvc) \leq \E_{\muc} l(\rvc)^* < \Hc(\rvc) + 1$\footnote{The `+1' can be reduced by more \emph{clever} lossless compression strategy - out of scope of this paper.}.
To minimize the rate we therefore minimize the entropy of the discrete code $\Hc(\rvc)$ so that 
\begin{align}\label{eq:loss}
\Ls := \E_{\mux} d(\rvx, \hat{\rvx}) + \reg \Hc(\rvc) \enspace .
\end{align}

\section{Quantization}\label{sec:quantization}

We employ the soft relaxation approach to quantization proposed in \citet{agustssonSofttoHardVectorQuantization2017} simplified similarly to \citet{mentzerConditionalProbabilityModels2018}.
However, instead of the scalar version \citet{mentzerConditionalProbabilityModels2018,habibianVideoCompressionRateDistortion2019} we use the \textbf{vector formulation of the quantization} as in \citet{oordNeuralDiscreteRepresentation2017} in which the $k$ quantization centers are the learned embedding vectors $\{\ve^{(j)}\}_{j=1}^k, \ \ve_i \in \sR^m$, columns of the $m \times k$ embedding matrix $\mE = [\ve^{(1)}, \ldots, \ve^{(k)}]$.


The quantizer $\gQE$ first reshapes the transformed data\footnote{For notational simplicity we regard the data as $d$ dimensional vectors. In practice, these are often $(d_1 \times d_2)$ matrices or even higher order tensors $(d_1 \times \ldots \times d_t)$. $\vz$ can be seen simply as their flattened version with $d = \prod_i d_i$.} $\vz \in \sZ \subseteq \sR^{d}$ into a $m \times d/m$ matrix $\mZ = [\vz^{(1)}, \ldots, \vz^{(d/m)}]$, then finds for each column $\vz^{(i)} \in \sR^m$ its nearest embedding and replaces it by the embedding vector index to output the $d/m$ dimensional vector of discrete codes $\vc = \gQE(\vz)$
\begin{align}\label{eq:quantization}
\gQE: \quad \vzh^{(i)} = \argmin_{\ve^{(j)}} \Vert \vz^{(i)} - \ve^{(j)} \Vert, \qquad \evc^{(i)} = \{j : \vzh^{(i)} = \ve^{(j)} \}, \quad i = 1, \ldots, d/m \enspace .
\end{align}
After transmission, the quantized latent representation $\hat{\mZ} = [\vzh^{(1)}, \ldots, \vzh^{(d/m)}]$ is recovered from $\vc$ by indexing to the shared codebook $\mE$ and decoded $\vxh = \gDp(\vzh)$, $\vzh = \mathrm{flatten}(\hat{\mZ})$. 
In practice, the quantized latent $\vzh$ can be used directly by the decoder at training in the \textbf{forward pass} without triggering the $\vc$ indexing operation.

The finite quantization operation in \eqref{eq:quantization} is non-differentiable.
To allow for the flow of gradients back to $\gEt$ and $\gQE$ we use a differentiable soft relaxation for the \textbf{backward pass}
\begin{align}\label{eq:soft_quantization}
\vzt^{(i)} = \sum_j^k \ve^{(j)} \softmax(-\sigma \Vert \vz^{(i)} - \ve^{(j)} \Vert) = \sum_j^k \ve^{(j)} \frac{\exp(-\sigma \Vert \vz^{(i)} - \ve^{(j)} \Vert)}{\sum_j^k \exp(-\sigma \Vert \vz^{(i)} - \ve^{(j)} \Vert)} \enspace ,
\end{align}
where instead of the hard encoding $\hat{\vzb}^{(i)}$ picking the single nearest embedding vector, the soft $\vzt^{(i)}$ is a linear combination of the embeddings weighted by their (softmaxed) distances.
The distortion loss is thus formulated as $d(\vx, \vxh) = d(\vx, \gDp[\mathrm{sg}(\vzh - \vzt) + \vzt])$, where $\mathrm{sg}$ is the stopgradient operator. 

The hard/soft strategy is different from the approach of \citet{oordNeuralDiscreteRepresentation2017} where they use a form of straight-through gradient estimator and dedicated codebook and commitment terms in the loss to train the embeddings $\mE$.
This is also different from \citet{williamsHierarchicalQuantizedAutoencoders2020}, where they use the relaxed formulation of \eqref{eq:soft_quantization} for both forward and backward passes in a fully stochastic quantization scheme aimed at preventing the mode-dropping effect of a deterministic approximate posterior in a hierarchical vector-quantized VAE.

\section{Minimizing the code cross-entropy}

Though the optimal lossless encoding is decided by the self-information of the code $-\log \pc(\vc)$, it cannot be used directly since $\pc$ is unknown.
Instead we replace the unknown $\pc$ by its estimated approximation
$\qc$, derive the code length $\hat{l}(\vc)$ from $\hat{i}_c = - \log \qc(\vc)$, and therefore minimize the expected approximate self-information, the cross-entropy $\cHc(\rvc) = - \E_{\muc} \log \qc(\rvc)$.
This, however, yields inefficiencies as $\cHc(\rvc) \geq \Hc(\rvc)$ due to the decomposition
\begin{align}\label{eq:crossentropy_KL}
\cHc(\rvc) = \KL(\pc \Vert \qc) + \Hc(\rvc) \enspace ,
\end{align}
where $\KL \geq 0$ is the Kullback-Leibler divergence between $\pc$ and $\qc$ which can be interpreted as the expected additional bits over the optimal rate $\Hc(\vc)$ caused by using $\qc$ instead of the true $\pc$.
In addition to $\gEt$, $\gDp$ and $\gQE$ we shall now therefore train also a probability estimator $\gPp : \{\vc\}_i^n \to \qc$ by minimizing the cross-entropy $\cHc(\rvc)$ so that the estimated $\qc$ is as close as possible to the true $\pc$, the $\KL$ is small, and the above mentioned inefficiencies disappear.

As we cannot evaluate the cross-entropy over the unknown $\muc$, we instead learn $\gPp$ by its empirical estimate over the sample data which is equivalent to minimizing the negative log likelihood (NLL)
\begin{align}\label{eq:empirical_crossentropy}
\argmin_\psi - \frac{1}{n}\sum_i^n \log \qc(\rvc = \vc_i), \quad \vc_i \sim \muc \enspace .
\end{align}
Similar strategy has been used for example in \citet{theisLossyImageCompression2017} and \citet{balleEndtoendOptimizedImage2017} both using some form of continuous relaxation of $\qc$ as well as in \citet{mentzerConditionalProbabilityModels2018} using an autoregressive PixelCNN as $\gPp$ to model $\qc(\vc) = \prod_i \qc(\evc_i | \evc_{i-1}, \ldots. \evc_1)$.

There is one caveat to the above approach.
In the minimization in \eqref{eq:empirical_crossentropy} the sampling  distribution $\mu_c$ is treated as fixed.
Minimizing the cross-entropy in such a regime minimizes the $\KL$ and hence the additional bits due to $\qc \neq \pc$ but not the entropy $\Hc(\rvc)$ (see \eqref{eq:crossentropy_KL}) which is treated as fixed and therefore not optimized for low rate as explained in section \ref{sec:quantization}.

This may seem natural and even inevitable since the samples $\vc$ are the result of sampling the data $\vx$ from the unknown yet fixed distribution $\mu_x$.
Yet, \textbf{the distribution $\mu_c$ is not fixed}.
It is determined by the learned transformation $\gTEt = \gQE \circ \gEt$ as the push-forward measure of $\mu_x$
\begin{align}
\mu_c[\rvc \in \mA] = \mu_c[\gTEt(\rvx) \in \mA] = \mu_x[\rvx \in \gTEt^{-1}(\mA)] \enspace ,
\end{align}
where $\mA \in \sC$ and $\gT^{-1}$ is the inverse image\footnote{The notation $\gT^{-1}$ here should not be mistaken for an \emph{inverse function} as $\gT$ is generally not invertible.} defined as $\gT^{-1}(\mA) := \{\rvx \in \sX : \gT(\rvx) \in \mA\}$.
Changing the parameters of the encoder $\gEt$ and the embeddings $\mE$ will change the measure $\mu_c$ and hence the entropy $\Hc(\rvc)$ and the cross-entropy $\cHc(\rvc)$ even with the approximation $\qc$ fixed.

We therefore propose to optimise the encoder and the embeddings so as to minimize the cross-entropy not only through learning better approximation $\qc$ but also through changing $\mu_c$ to achieve overall lower rate.
Since the discrete sampling operation from $\mu_c$ is non-differentiable, we propose to use a simple continuous soft relaxation similar the one described in section \ref{sec:quantization}.
Instead of using the deterministic non-differentiable code assignments
\begin{align}\label{eq:hardprobs}
\pc(\rvc^{(i)} = j) =
\begin{cases}
1 & \text{if} \ \vzh^{(i)} = \ve^{(j)} \\
0 & \text{otherwise}
\end{cases},
\quad i = 1, \ldots, d/m
\end{align}
we use the differentiable soft relaxation
\begin{align}\label{eq:softprobs}
\hat{p}_c(\rvc^{(i)} = j) = \softmax(-\sigma \Vert \vz^{(i)} - \ve^{(j)} \Vert) = \frac{\exp(-\sigma \Vert \vz^{(i)} - \ve^{(j)} \Vert)}{\sum_j^k \exp(-\sigma \Vert \vz^{(i)} - \ve^{(j)} \Vert)} \enspace .
\end{align}

Our final objective is the minimization of the empirical loss composed of three terms: the distortion, the \emph{soft} cross-entropy, and the \emph{hard} cross-entropy
\begin{align}\label{eq:loss_final}
\Lsh(\theta, \mE, \phi, \psi) := \frac{1}{n}\sum_i^n d(\vx_i, \vxh_i) + \alpha \, s(\vc_i) + \beta \, h(\vc_i) \enspace .
\end{align}
The distortion may be the squared error $d(\vx, \vxh) = \Vert \vx - \vxh \Vert^2$ or other application-dependent metric (e.g. multi-scale structural similarity index for visual perception in images).
Through the distortion we optimize the parameters of the decoder $\gDp$ and using the relaxation described in section \ref{sec:quantization} for the backward pass also the parameters of the encoder $\gEt$ and the quantization embeddings $\gQE$.

The \emph{hard} cross-entropy loss is
\begin{align}
h(\vc) = - \frac{m}{d} \sum_j^{d/m} \sum_{\vc^{(j)} = 1}^k \pc(\vc^{(j)}) \log \qc(\vc^{(j)})
= - \frac{m}{d} \sum_j^{d/m} \log \qc(\vc^{(j)}) \enspace ,
\end{align}
where we treat the dimensions of the vector $\vc = [\vc^{(1)}, \ldots, \vc^{(d/m)}]$ as independent so that the approximation $\qc$ can be used directly as the entropy model for the lossless arithmetic coding (expects the elements of the messages to be sampled i.i.d. from a single distribution).
The loss simplifies to the final form due to the 0/1 probabilities of the deterministic code-assignments in \eqref{eq:hardprobs}.
Through the hard cross-entropy loss we learn the parameters of the probability model $\gPp$ outputting the $\qc$ distribution.

The \emph{soft} cross-entropy loss is
\begin{align}
s(\vc) = - \frac{m}{d} \sum_j^{d/m} \sum_{\vc^{(j)} = 1}^k \hat{p}_c(\vc^{(j)}) \log \textrm{sg}[\qc(\vc^{(j)})] \enspace ,
\end{align}
which uses the differentiable soft relaxation $\hat{p}_c$ of \eqref{eq:softprobs}. 
This allows for back-propagating the gradients to the encoder $\gEt$ and the quantizer $\gQE$. 
We use the $sg$ operator here to treat $\qc$ as fixed in this part of the loss preventing further updating of the parameters of the probability model $\gPp$.

\section{Experiments}

As a proof of concept we conducted a set of experiments on the tiny 32x32 CIFAR-10 images \citep{krizhevskyLearningMultipleLayers2009}.
We use similar architecture of the encoder $\gEt$ and decoder $\gDp$ as \citet{mentzerConditionalProbabilityModels2018} (without the spatial importance mapping), with the downsampling and upsampling stride-2 convolutions with kernels of size 4, and 10 residual blocks with skip connections between every 3.
We fix the annealing parameter $\sigma=1$ and the loss hyper-parameter $\beta=1$\footnote{In our preliminary experiments the results were not very sensitive to $\beta$. In fact, $\beta$ influences only the speed with which the probability model $\gPp$ is trained compared to the other components of the model updated through the other parts of the loss.}.
We use ADAM with default pytorch parameters, one cycle cosine learning rate and train for 15 epochs.
The code is available at:
\url{https://bitbucket.org/dmmlgeneva/softvqae/}.

We first compare the vector quantization (VQ) approach where the codebook is composed of $m$-long vectors versus the scalar (SQ) approach where it contains scalars $m=1$ as e.g. in \cite{mentzerConditionalProbabilityModels2018}.
By construction the VQ version needs to transmit shorter messages for the same level of downsampling. For example, with 8-fold downsampling to $4 \times 4$ latents $\vz$ with 8 channels the discrete codes $\vc$ of SQ have $d/m = 128$ elements.
In VQ, the channel dimension forms the rows of the matrix $\mZ$ with $m=8$ and the $\vc$ messages to be encoded and transmitted have only $d/m = 16$ elements.
On the other hand, in the scalar version each of the 8 channels is represented by its own code and therefore allows for more flexibility compared to a single code for the whole vector.

\begin{figure}[h]
\begin{center}
\includegraphics[scale=0.5]{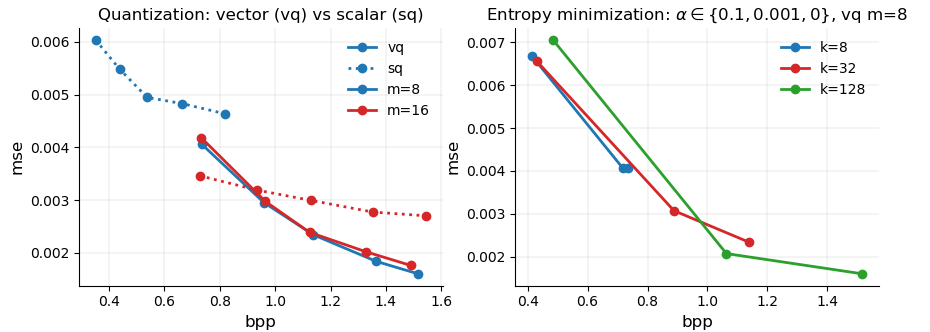}
\end{center}
\caption{Rate-distortion curves. Rate is expressed in bits per pixel (bpp) of the original images, distortion is expressed in the mean squared error (mse) between the original and reconstructed images. See text for description.}\label{fig:plots}
\end{figure}

Our preliminary experiments confirm the superiority of the VQ. In figure \ref{fig:plots} left we plot the rate-distortion for comparable parts of the trade-off space.
The VQ models use 2-folds downsampling resulting in $d/m = 16 \times 16 = 256$ long messages $\vc$.
The two curves are for embeddings (latent $\vz$ channels) with size $m=8$ and $m=16$ respectively and the points around the curves are the result of increasing the size of the dictionary $\mE$ as $k = [8, 16, 32, 64, 128]$ from left to right.
The SQ models use 8-folds downsampling with latent $\vz$ channels 8 and 16 resulting in $d/m = 4 \times 4 \times 8 = 128$ and $d/m = 4 \times 4 \times 8 = 256$ ($m=1$ here). We observe that VQ clearly achieves better trade-offs being in the bottom-left of the plots.

We next confirm the effectiveness of the soft cross-entropy term $s(\vc)$ in our final loss formulation in \eqref{eq:loss_final}.
Increasing the $\alpha$ hyper-parameter should put more importance on the rate minimization (through the entropy) as compared to the distortion.
In the right plot of figure \ref{fig:plots} we compare the points $k=[8, 32, 128]$ of the \emph{`vq, 8'} curve for values $\alpha = [0.01, 0.001, 0]$ from left to right.
With the highest $\alpha = 0.01$, the objective trade-off searches for the lowest rate tolerating higher distortion. The lower the $\alpha$, the less we push for low rates which allows for smaller distortion. 
This behaviour corresponds well to the expected and desirable one where, as formulated in \eqref{eq:trade_off}, we can now directly control the trade-off between the two competing objectives by setting the hyper-parameter $\alpha$.

In the appendix we provide examples of the learned $\qc$ histograms for \emph{`vq, 8, k=32'} with different values of $\alpha$ showing the concentration of the measure into a few points in the support for high $\alpha$.


\bibliography{softvqae}
\bibliographystyle{iclr2021_conference}

\appendix
\section{Appendix}

Histograms $\qc$ learned by the probability model $\gPp$ for the \emph{`vq, 8'} models with the number of embedding vectors $k=32$ $\approx$ the red line in the right graph in Figure \ref{fig:plots} with increasing $\alpha \in \{0, 0.001, 0.01\}$.
For the highest $\alpha = 0.01$ the distribution is concentrated into a few points in the support resulting in the lowest entropy (and therefore the best rate) but highest distortion.

\begin{figure}[h]
\begin{center}
\includegraphics[scale=0.5]{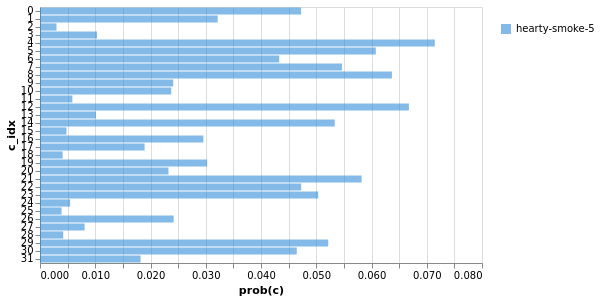}
\end{center}
\caption{Learned historgram $\qc$ for $\alpha = 0$}
\end{figure}
\begin{figure}[h]
\begin{center}
\includegraphics[scale=0.5]{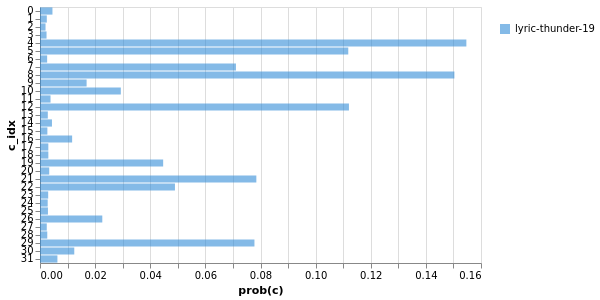}
\end{center}
\caption{Learned historgram $\qc$ for $\alpha = 0.001$}
\end{figure}
\begin{figure}[h]
\begin{center}
\includegraphics[scale=0.5]{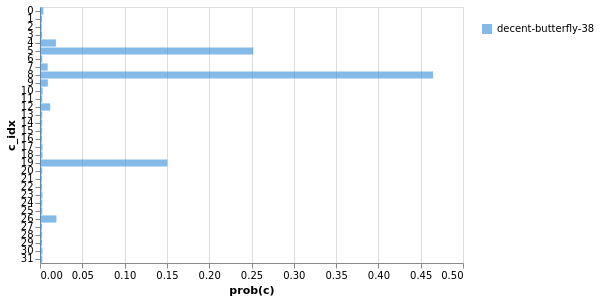}
\end{center}
\caption{Learned historgram $\qc$ for $\alpha = 0.01$}
\end{figure}

\end{document}